%% file: main.tex
\documentclass[journal]{IEEEtran}

\usepackage{amsmath,amsfonts}
\usepackage{array}
\usepackage{float}
\usepackage[utf8]{inputenc}
\usepackage[table]{xcolor}
\usepackage[linesnumbered,ruled]{algorithm2e}
\usepackage[left=0.625in, right=0.625in, top=0.75in, bottom=1in]{geometry}

\usepackage{booktabs}
\usepackage[english]{babel}

\usepackage{amsmath, amssymb, amsfonts, amsthm,algpseudocode}
\usepackage{diagbox} 
\usepackage{enumitem}
\usepackage{forest}
\usepackage{geometry, graphicx}
\usepackage{hyperref}
\usepackage{longtable}
\usepackage{multirow, makecell}
\usepackage{newunicodechar}
\usepackage{pgf}
\usepackage{subcaption}
\usepackage{caption}
\usepackage{tabularx, tikz}
\usepackage{verbatim}

\newunicodechar{​}{}
\usepackage{balance}

\definecolor{pink}{RGB}{244, 183, 190} 
\definecolor{deep_pink}{RGB}{239,148,159}
\definecolor{light_green}{RGB}{227,242,217} 
\definecolor{blue}{RGB}{208,228,246} 
\definecolor{deep_orange}{RGB}{255,136,16}
\definecolor{orange}{RGB}{244,179,130} 
\definecolor{blue_green}{RGB}{210,244,242} 
\definecolor{light_yellow}{RGB}{254,231,150} 
\definecolor{light_gray}{RGB}{166,166,166}
\usepackage{placeins}
\def\BibTeX{{\rm B\kern-.05em{\sc i\kern-.025em b}\kern-.08em
    T\kern-.1667em\lower.7ex\hbox{E}\kern-.125emX}}
    
\title{Semantic-preserved Augmentation with Confidence-weighted Fine-tuning for Aspect Category Sentiment Analysis}

\author{Yaping Chai, Haoran Xie, Joe S. Qin
\thanks{This work was supported by a grant from the Research Grants Council of the Hong Kong Special Administrative Region, China (R1015-23), and the Faculty Research Grant (SDS24A8) and the Direct Grant (DR25E8) of Lingnan University, Hong Kong. \emph{(Corresponding author: Haoran Xie.)}}

\thanks{Yaping Chai, Haoran Xie, and Joe S. Qin are with the School of Data Science, Lingnan University, Hong Kong (e-mail: yapingchai@ln.hk; hrxie@ln.edu.hk; joeqin@ln.edu.hk).}
}
\begin{document}
\maketitle
\begin{abstract}
Large language model (LLM) is an effective approach to addressing data scarcity in low-resource scenarios. Recent existing research designs hand-crafted prompts to guide LLM for data augmentation. We introduce a data augmentation strategy for the aspect category sentiment analysis (ACSA) task that preserves the original sentence semantics and has linguistic diversity, specifically by providing a structured prompt template for an LLM to generate predefined content. In addition, we employ a post-processing technique to further ensure semantic consistency between the generated sentence and the original sentence. The augmented data increases the semantic coverage of the training distribution, enabling the model better to understand the relationship between aspect categories and sentiment polarities, enhancing its inference capabilities. Furthermore, we propose a confidence-weighted fine-tuning strategy to encourage the model to generate more confident and accurate sentiment polarity predictions. Compared with powerful and recent works, our method consistently achieves the best performance on four benchmark datasets over all baselines.

\end{abstract}


\input{1_introduction}

\input{2_related_work}

\input{3_method}

\input{4_experiment}

\input{5_conclusion}
\section*{Acknowledgment}
The research described in this article has been supported by a grant from the Research Grants Council of the Hong Kong Special Administrative Region, China (R1015-23), and the Faculty Research Grant (SDS24A8) and the Direct Grant (DR25E8) of Lingnan University, Hong Kong.

\bibliographystyle{IEEEtran}
\bibliography{main}
\balance

\end{document}

%% file: 1_introduction.tex
\section{Introduction}

\begin{figure*}[htbp]
    \centering
    \includegraphics[width=0.70\textwidth]{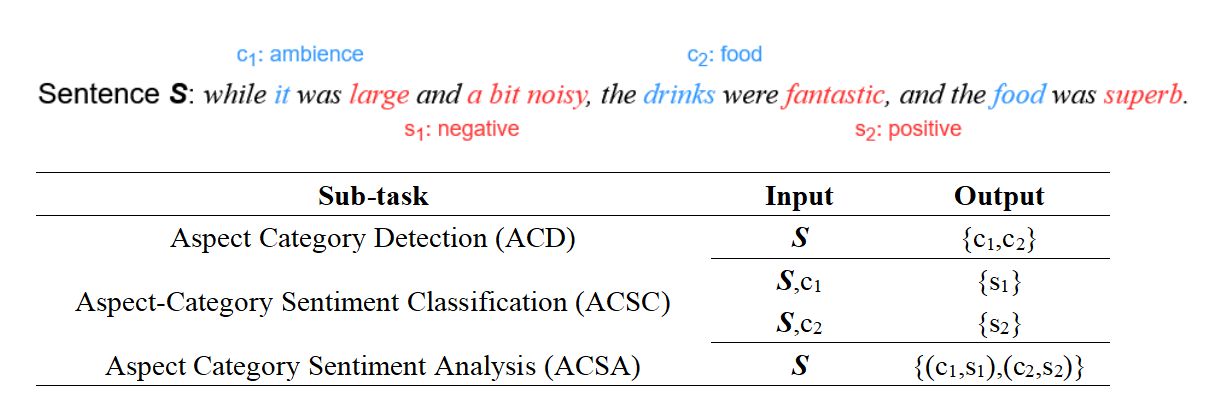}
    \caption{An illustration of ACD, ACSC and ACSA subtasks. The c, s, and \textbf{\textit{$ S $}} denote aspect category, sentiment polarity, and input sentence, respectively.}
    \label{fig:ACSA}
\end{figure*}

Aspect-based sentiment analysis (ABSA) has received significant attention from researchers in natural language processing (NLP) for its ability to provide finer-grained expression of sentiments \cite{DL-absa, add1,add2}. ABSA aims to identify specific aspects of an entity and determine the sentiment polarity associated with each aspect. This granularity level is crucial for applications such as e-commerce, social media, and customer feedback analysis. ABSA includes multiple subtasks, three of which are aspect category detection (ACD), aspect-category sentiment classification (ACSC) and aspect category sentiment analysis (ACSA). ACD involves identifying predefined aspect categories mentioned in a given text. ACSC extends ACD by assigning a sentiment polarity to each detected aspect category. While ACSA is an end-to-end task that combines both ACD and ACSC, which jointly predicts both the aspect categories and their corresponding sentiment polarities. Figure \ref{fig:ACSA} demonstrates an example for each of the three subtasks.

One challenge of ABSA is the scarcity of annotated high-quality training data \cite{cmlm,SPDAug-ABSA,rsda, DS2-ABSA}. However, manual annotation is time-consuming and requires significant human resources. Since large language models (LLMs) have demonstrated remarkable text generation ability, more research has begun to employ LLMs for data augmentation to solve ABSA tasks in low-resourced scenarios. The recent works \cite{rsda,DS2-ABSA} on ABSA data augmentation have generated a large amount of data, but they do not guarantee semantic consistency between the synthetic sentences and the original sentences. Other works, such as \cite{SPDAug-ABSA,cmlm} obtain augmented data by replacing words, however they largely preserve the original sentence structure and limit the expression diversity of the generated data. Additionally, \cite{cmlm} introduces \textit{semantic consistency} and finds that the generated dataset that preserves the semantics of labelled data is beneficial for improving the model’s performance. 

Based on the above research findings, we aim to address the data scarcity issue in ACSA tasks. Concretely, we design specific prompt templates to guide LLMs to generate data that is significantly different from the original text while preserving its semantics. To ensure the quality of the synthetic data, we utilise a post-processing technique to filter out generated data that deviates too much from the original sentence's semantics. Moreover, fine-tuning is an effective approach to adapting pre-trained models to downstream tasks while retaining general knowledge from the pre-training phase. We propose a confidence-weighted fine-tuning approach, whose key idea is to give additional weights based on the confidence level of the predictions when the model makes the correct predictions, thereby encouraging the model to make more confident and accurate predictions.

In summary, the main contributions of our work are as follows:
\begin{itemize}
\item We introduce a novel data augmentation method that leverages LLMs to preserve the original sentence semantics while introducing linguistic diversity. 
\item We propose a confidence-weighted fine-tuning strategy that encourages the model to make more confident and accurate predictions.
\item We conduct a comprehensive analysis of key hyperparameter variations in our experimental setup. Our proposed method consistently achieves the best performance in all four benchmark datasets on both the ACSC and ACSA tasks. 

\end{itemize}

%% file: 2_related_work.tex
\section{Related Work}
\label{sec-Preliminaries}

\subsection{Aspect Category Sentiment Analysis}
Early research considered ACSA as a classification task. AAGCN \cite{AAGCN} obtains the final sentiment classification result based on the probability distribution output through the softmax layer. AC-MIMLLN \cite{AC-MIMLLN} classifies the sentiments of aspect categories involved in a sentence by aggregating the sentiments of key instances. PBJM \cite{PBJM} regards sentiment polarity prediction as a binary classification task and utilises natural language prompt templates to perform aspect category detection (ACD) and aspect-category sentiment classification (ACSC) tasks.

To overcome the limitations of traditional classification, which cannot directly leverage pre-trained knowledge and cannot generalise better to new domains, BART-generation \cite{ACSA-gen} first proposes treating ACSA as a text-generation task. It can better utilise the advantages of BART \cite{bart} in the input’s semantic level summarisation. PyACSA \cite{pyacsa} develops an ACSA framework and uses Flan-T5 \cite{flan-t5} as the generative model, converting the ACSA task into a text-to-text format. 

Recent researchers have begun to explore the capabilities of LLMs in sentiment analysis and use them to unify various subtasks of ABSA. LEGO-ABSA \cite{LEGO-ABSA} proposes to solve diverse ABSA tasks in a unified generative framework by controlling task prompt types composed of multiple elements. ChatABSA \cite{chatabsa} is a unified framework that can convert five complex subtasks into the prompt format.

\subsection{Prompt-based Techniques}

Prompt engineering, as a technique for guiding language models to output expected content, has undergone significant changes since the introduction of early pre-trained models like BERT \cite{BERT}, which features cloze-style prompting and influences later developments in fill-in-the-blank tasks. 

Along with the magnificent breakthrough of LLMs, prompt-based learning has become a revolutionary technique for expanding the capabilities of LLMs. \cite{gpt-3} has demonstrated that zero-shot and few-shot prompting significantly impact model performance.

Technologies such as Chain-of-Thought prompting \cite{cot} and self-consistency prompting \cite{self-consistency} further enhance LLMs’ reasoning ability by explicitly constructing multiple steps. In the ABSA task, RVISA \cite{rvisa} solves implicit sentiment analysis by providing reasoning prompting for LLMs to obtain convincing rationales. 

\subsection{Data Augmentation}

The earliest widely used data augmentation methods include replacing synonyms, shuffling word order, and randomly deleting words \cite{da}. 

With the flourishing development of LLMs, many studies have leveraged prompt-based techniques for data augmentation, which solves the problem of data scarcity by designing crafted prompts. Among them, in the ABSA task, RSDA \cite{rsda} performs well in cross-domain tasks but does not consider whether the semantic deviation of the generated sentence was too large compared to the original sentence. SPDAug-ABSA \cite{SPDAug-ABSA} proposes to preserve the semantics of the original sentence by considering each word’s importance in the text sequence. Then, they use two replacement strategies to replace unimportant tokens. However, this is still a token-level transformation that cannot bring diversity to the entire sentence. 

To address the above limitations of data augmentation in sentiment analysis, we utilise the exceptional generation capability of LLMs to produce sentences that are rich in expression while ensuring semantic consistency with the original sentence.

%% file: 3_method.tex
\section{Method}

\begin{figure}[htbp]
    \centering
    \includegraphics[width=0.5\textwidth]{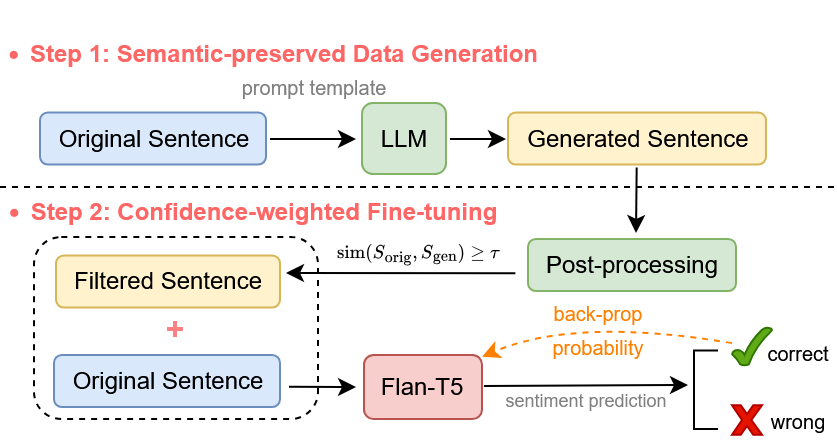}
    \caption{The overview of our method, including data augmentation, post-processing, and confidence-weighted fine-tuning.}
    \label{fig:ltm}
\end{figure}

Aiming to obtain sentences that are semantically consistent with the original sentence and have expression differences through data augmentation, we preserve the categories and their corresponding sentiment polarities of the original sentences, design a prompt template for the ACSA task and guide the LLM to generate sentences based on the predefined prompt template. To ensure data quality, we use a post-processing mechanism to filter out samples that deviate significantly from the original sentence semantics. Finally, we acquire a synthetic dataset that contains category-polarity pairs. Moreover, fine-tuning creates a bridge between general pre-trained models and the unique requirements of specific applications. We propose a novel fine-tuning strategy that introduces a confidence-weighted loss function. The key idea is to provide additional weights based on the confidence of the prediction when the model makes a correct prediction, thereby encouraging the model to make more confident and accurate predictions. The overall architecture is shown in Figure \ref{fig:ltm}.

\subsection{Task Definition}
Aspect category detection (ACD) aims to identify predefined aspect categories associated with a given text \cite{ACSA-gen}. Unlike aspect term extraction, which identifies explicitly mentioned aspect terms, ACD focuses on categorising the implicit or explicit aspects discussed in the text. The categories are predefined and domain-specific. Given an input sentence \textbf{\textit{$ S $}}, the goal is to predict a set of aspect categories $ C $, where \(C = \{c_i\}_{i = 1}^m\) is a category set composed of all $c_i$, $c_i$ represents a category in the sentence and $m$ is the total number of all categories in the sentence.

Aspect-category sentiment classification (ACSC) extends ACD by assigning a sentiment polarity $s_i$ to each detected aspect category $c_i$. The sentiment polarity can be categorised into three main classes: positive, neutral, and negative.

Aspect category sentiment analysis (ACSA) is the end-to-end task that combines both ACD and ACSC. It aims to jointly predict both the aspect categories and their corresponding sentiment polarities. Given an input sentence \textbf{\textit{$ S $}}, the objective is to predict a set of category-sentiment pairs \(\{c_i,s_i\}_{i = 1}^m\). 

\subsection{Semantic-preserved Augmentation}
\label{da}
\subsubsection{Augmented Sentences Generation}
\begin{figure}[htbp]
    \centering
    \includegraphics[width=0.45\textwidth]{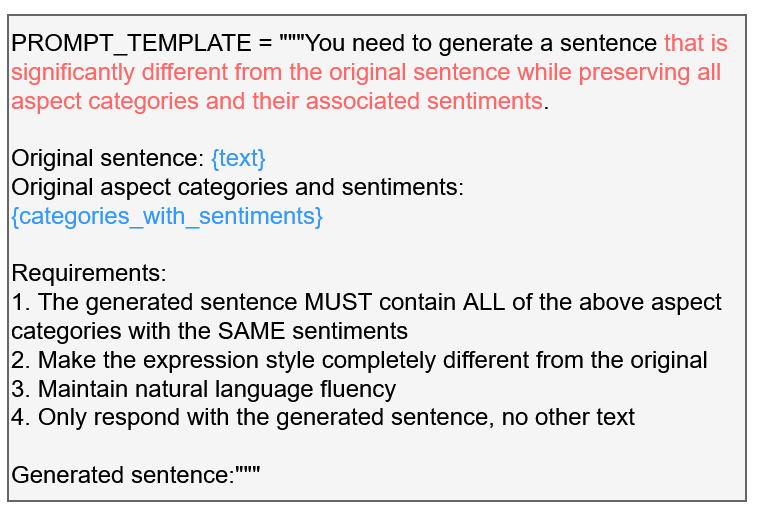}
    \caption{Structured prompt template for generating sentences.}
    \label{fig:sentence}
\end{figure}

One of the challenges of ACSA is data scarcity, especially due to the need for fine-grained annotation of sentences across multiple categories and their related sentiments. However, collecting and annotating large-scale high-quality datasets is both expensive and time-consuming \cite{da}. Traditional data augmentation methods, such as synonym replacement and back translation, cannot maintain the semantics of the original sentence or lack linguistic diversity. To address these limitations, we utilise LLMs for text data augmentation to generate diverse and semantically consistent sentences. The structured prompt template for sentence generation is shown in Figure \ref{fig:sentence}.

Given a labeled sample \(x = ({S_{orig}, y) }\), where \(S_{orig}\) denotes the original sentence and \(y = {(c_i, s_i)}_{i=1}^n\) is a set of \(n\) category-sentiment pairs with aspect category \(c_i\) and sentiment polarity \(s_i\). We preserve all \( {(c_i, s_i)}_{i=1}^n \in y\) of the original sentence and explicitly guide the model to produce a synthetic sentence \( S_{gen} \) that is rich in expression and retain the meaning of the original sentence in the constructed prompt template. Pair \( S_{gen} \) generated by LLM with the original label \( y\) to form an augmented sample \(x' = ({S_{gen}, y)}\). Finally, we obtain a set of generated sentences with their category-polarity pairs.

\subsubsection{Semantic Consistency Filtering }

The data generated by LLMs cannot fully guarantee high quality, and adopting a refinement strategy for the augmented data is beneficial for model training \cite{da}. To further ensure the quality of the generated dataset, we employ a post-processing technique to filter out sentences with significant semantic deviations from the original sentences. Specifically, we leverage Sentence-BERT (SBERT) \cite{sbert}, a sentence embedding model, to measure the semantic similarity between the original and synthetic sentences. SBERT maps sentences into a dense, high-dimensional vector space where semantically similar sentences are positioned closer together. Given an original sentence \( S_{\text{orig}} \) and a generated sentence \( S_{\text{gen}} \), SBERT computes their embeddings as follows: $v_{\text{orig}} = \text{SBERT}(S_{\text{orig}}),\quad v_{\text{gen}} = \text{SBERT}(S_{\text{gen}})$. These vector representations capture not only the lexical information but also deeper contextual and semantic properties of the sentences. 

We calculate their cosine similarity to quantify the semantic consistency between the original and generated sentences:  
\[
\text{sim}(S_{\text{orig}}, S_{\text{gen}}) = \frac{v_{\text{orig}} \cdot v_{\text{gen}}}{\|v_{\text{orig}}\| \|v_{\text{gen}}\|}
\]
To eliminate semantically inconsistent and low-quality sentences, we define a similarity threshold \( \tau \). If the similarity score falls below \( \tau \), the generated sentence is considered too semantically distant from the original and is discarded:  
\[
S_{\text{gen}} \text{ is retained if } \text{sim}(S_{\text{orig}}, S_{\text{gen}}) \geq \tau
\]
The threshold \( \tau \) is determined empirically based on validation experiments to ensure that we preserve diverse but semantically faithful sentences while filtering out overly deviating outputs. 


\subsection{Confidence-weighted Fine-tuning}

Fine-tuning can significantly affect the performance of NLP models, transforming general language models into specialised tools with enhanced capabilities for specific domains and tasks. We propose using a confidence-weighted approach to fine-tune the model. 

Given a training sample \( x_i \), \( y_i \) represents the ground-truth label for sample \( x_i \) and \( p_i^{(y_i)} \) denotes the predicted probability assigned to the \( y_i \). The standard cross-entropy loss for sample $x_i$ is given by:
\begin{equation}
\mathcal{L}_\text{CE}(y_i) = -\log p_i^{(y_i)}
\end{equation}

To enhance the model's ability to generate confident and accurate sentiment predictions, we propose a confidence-weighted fine-tuning strategy which integrates a confidence value based on prediction correctness and probability, allowing the model to learn more confident and accurate predictions.

The confidence value ensures that the correct predictions contribute to model training based on their confidence level. Incorrect predictions do not contribute to confidence, using standard cross-entropy loss to train the model. The confidence value $v_i$ is represented by:

\begin{equation}
v_i = 
\begin{cases} 
\max_{c} \, p_i^{(c)}, & \hat{y}_i = y_i\\
0, & \hat{y}_i \neq y_i
\end{cases}
\end{equation}
where \( \hat{y}_i \) is the predicted sentiment polarity for sample \( i \), $p_i^{(c)} $ represents the prediction probability of the model for the $i$-th sample on category $c$ and $ \max_{c} \, p_i^{(c)}$ represents the model's maximum predicted probability (i.e., its confidence). 

To incorporate the confidence value into the training process, we propose a confidence-weighted loss function, as shown in the following equation:

\begin{equation}
\mathcal{L}_i^\text{v} = \mathcal{L}_\text{CE}(y_i) \cdot (1 + \alpha \cdot v_i)
\end{equation}
where \( \alpha \in \mathcal{R} >0 \) is a hyperparameter controlling the influence of the confidence. When \(v_i>0 \), the confidence value affects the weight of the loss, thereby enhancing the correct prediction probability. When \(v_i=0 \), the loss becomes the standard cross-entropy loss.

The overall training objective over a dataset \( \mathcal{D} \) is computed as the average of the confidence-weighted losses:

\begin{equation}
\mathcal{L}_\text{v} = \frac{1}{|\mathcal{D}|} \sum_{i=1}^{|\mathcal{D}|} \mathcal{L}_i^\text{v}
\end{equation}

%% file: 4_experiment.tex
\section{Experiment}

\subsection{Datasets and Evaluation Metrics} 
In the experiment, we use four public datasets: Rest15 and Lap15 from the SemEval-2015 task 12 \cite{sem15}, and Rest16 and Lap16 from the SemEval-2016 task 5 \cite{sem16}. The statistics of each dataset are demonstrated in Table \ref{tab:dataset}. We evaluate the model using precision (P), recall (R), and F1 score (F1) for evaluation.

\input{table/dataset}

\input{table/ACSC}

\input{table/ACSA}

\subsection{Models and Baselines}
In the experiment, we utilise GPT-4o \footnote{\url{https://platform.openai.com/docs/models/}} as the generative model for data augmentation. Encoder-decoder models like Flan-T5-xl \cite{flan-t5} are pre-trained on massive corpora, giving them a stronger understanding of syntax and semantics. Moreover, Flan-T5-xl \cite{flan-t5} has been fine-tuned on a large set of tasks with natural language instructions, making it better handle implicit sentiment and domain-specific expressions. We employ Flan-T5-xl \cite{flan-t5} for fine-tuning. 

For the ACSC task, the parameters $\alpha$ we used for Rest15, Rest16, Lap15, and Lap16 datasets are 0.4673, 0.3045, 0.3045 and 0.6123, respectively. For the ACSA task, the parameters $\alpha$ we used are 0.1713, 0.5903, 0.2526 and 0.5378, which were tuned by employing Optuna \footnote{\url{https://github.com/optuna/optuna}}. In all experiments, the similarity threshold $\tau$ was 0.7. As we discuss the effect of $\tau$ on the model performance in Section \ref{tau}, when $\tau$ =0.7, our model achieves the best result in most of the benchmarks. To test the effectiveness of our model in ACSC and ACSA tasks, we compare it with multiple baselines according to different tasks. 

For the ACSC task, we compare it with the following methods: CAER-BERT \cite{CAER-BERT}, SCAN \cite{SCAN}, AC-MIMLLN \cite{AC-MIMLLN}, LC-BERT \cite{LC-BERT}, EDU-capsule \cite{EDU-Capsule}, Hier-BERT \cite{Hier-BERT}, Hier-GCN-BERT \cite{Hier-BERT}, MvP \cite{mvp}, ECAN \cite{ecan}. Among these baselines, ECAN \cite{ecan} exceeds other baselines in the ACSC task. 

For the ACSA task, we compare it with the following methods: Pipeline-BERT, Cartesian-BERT, Addonedim-BERT \cite{Addonedim-BERT}, AS-DATJM \cite{AS-DATJM}, Hier-BERT \cite{Hier-BERT}, Hier-Trans-BERT \cite{Hier-BERT}, Hier-GCN-BERT \cite{Hier-BERT}, PBJM \cite{PBJM}. Among them, PBJM \cite{PBJM} achieves the best performance in the ACSA task.

\subsection{Main Results}

Table \ref{tab:ACSC_result} and Table \ref{tab:ACSA_result} illustrate the performance of our proposed method with data augmentation (DA) and without data augmentation (w/o DA), compared to a set of baselines for the ACSC and the ACSA task on four benchmark datasets: Rest15, Rest16, Lap15, and Lap16. All baseline results are from ECAN \cite{ecan} and PBJM \cite{PBJM}, respectively, which are the best models among each task's baselines. Notably, both variants of our proposed method consistently outperforms all other baseline models across all datasets, achieving the highest experimental evaluation results.

\begin{figure*}[htbp]
    \centering
    \includegraphics[width=0.80\textwidth]{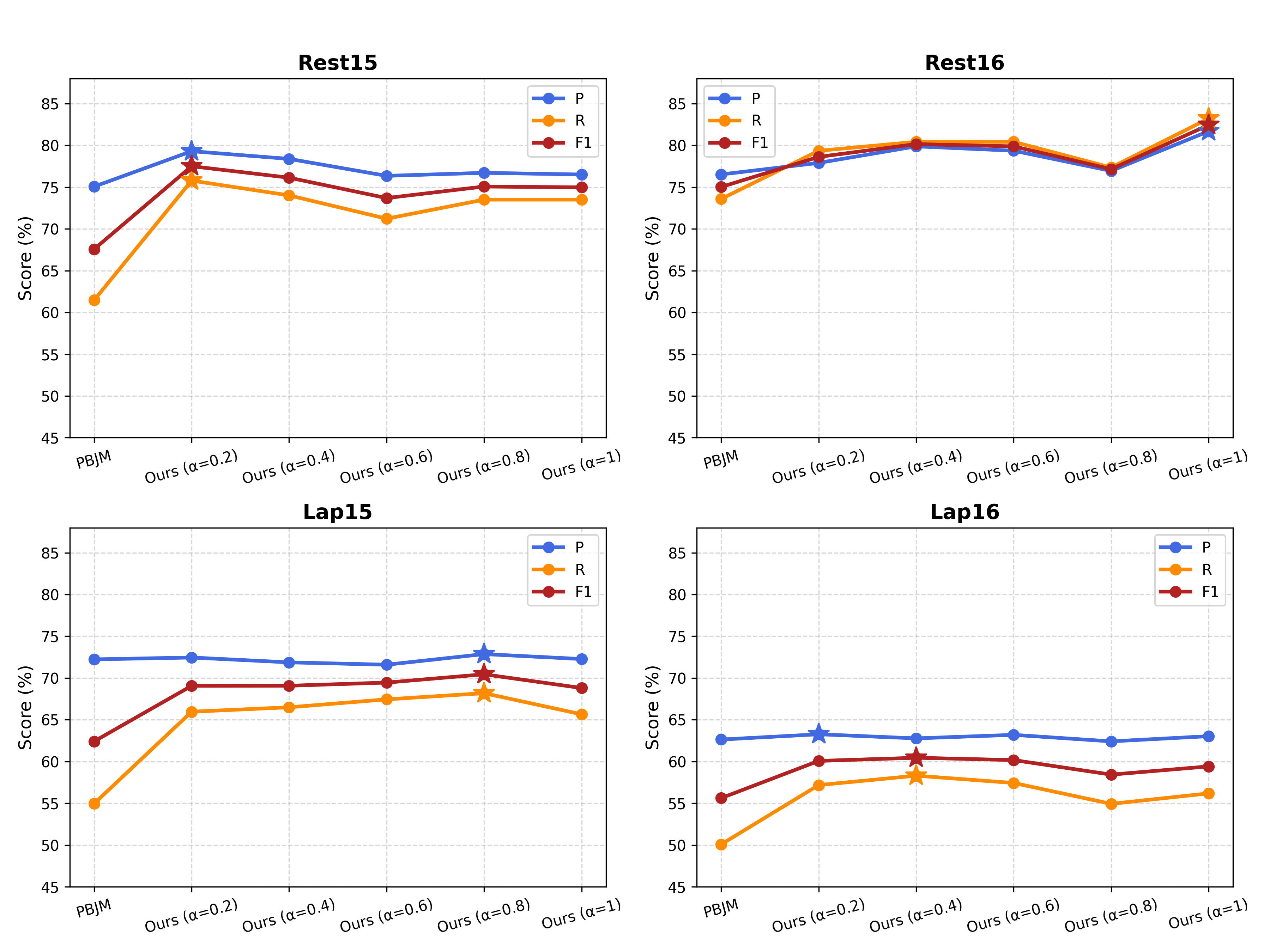}
    \caption{Precision, Recall, and F1 scores on four datasets with different values of $\alpha $ for the ACSA task. PBJM \cite{PBJM} is the baseline. The best metric values are highlighted with a star $\star$.}
    \label{fig:alpha}
\end{figure*}

\subsubsection{Performance of Aspect Category Sentiment Classification}

On the Rest15 dataset, our model without DA achieves the best overall performance, with an F1 score of 89.46\%, which is +10.37\% higher than ECAN. Notably, it also achieves higher Precision (89.75\% compared to 84.38\%) and Recall (90.53\% compared to 74.43\%). Although the performance of Ours (DA) has slightly decreases with an F1 of 86.72\%, it is still +7.63\% F1 higher than ECAN. This indicates that although data augmentation does not improve performance in this dataset, the core model architecture is still effective.

On the Rest16 dataset, the model with data augmentation achieves the highest F1 score of 93.44\%, significantly surpassing ECAN by +11.24\%. Both Precision (93.84\%) and Recall (93.83\%) reach very high results, indicating the effectiveness of data augmentation in the Rest16 dataset. Even without augmentation, Ours (w/o DA) has an F1 score of 91.88\%, which is +9.68\% higher than ECAN. Similarly, For the Lap15 and Lap16 datasets, our DA-enhanced model once again achieves the best F1 score, with remarkably high Precision and Recall. Our model without augmentation also significantly outperforms ECAN. These results underline the superiority of our method in both DA and w/o DA settings.

\begin{figure*}[htbp]
    \centering
    \includegraphics[width=0.80\textwidth]{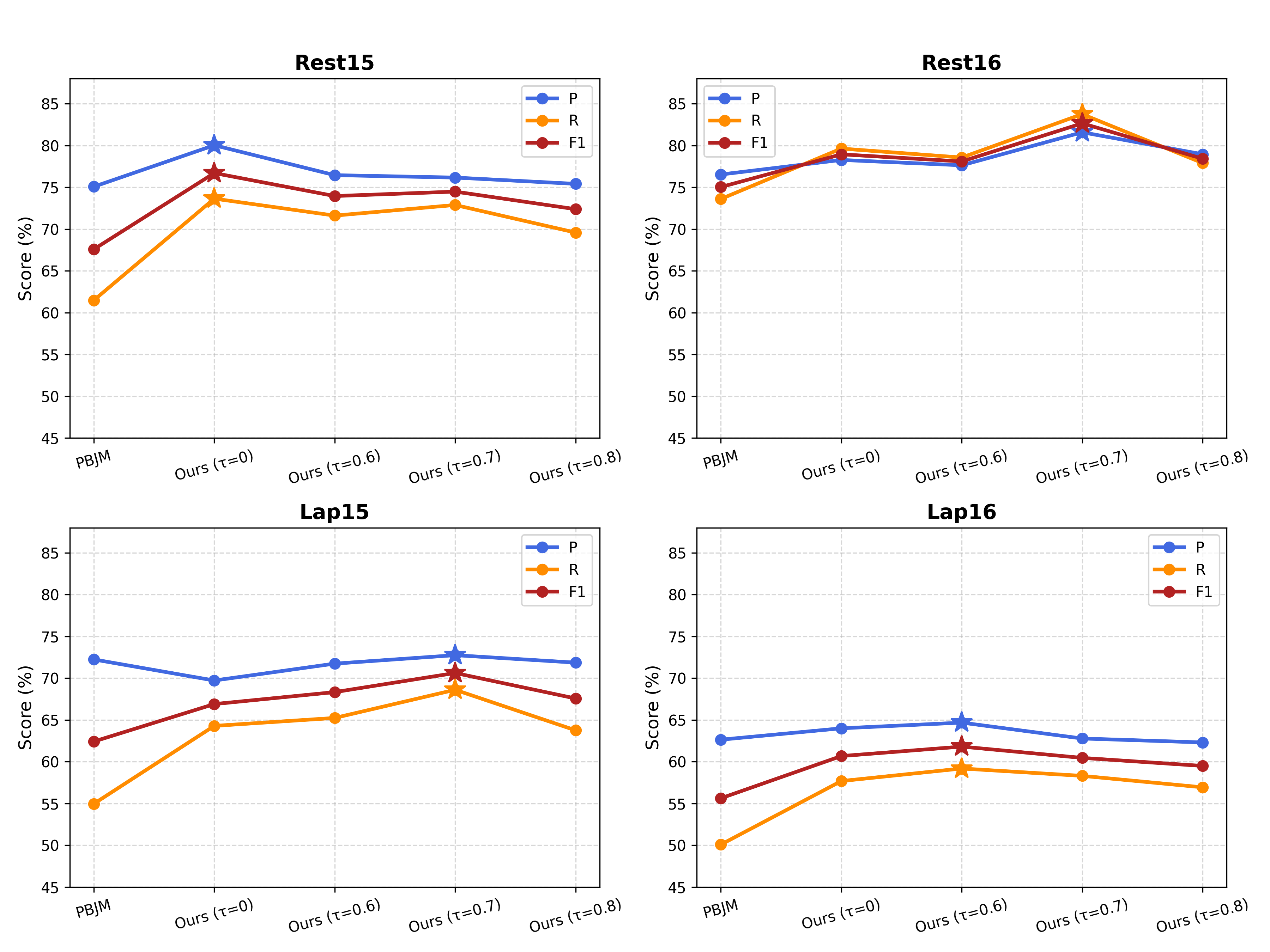}
    \caption{Precision, Recall, and F1 scores on four datasets with different values of the semantic consistency threshold $\tau$ for the ACSA task. PBJM \cite{PBJM} is the baseline, and $\tau=0$ indicates no filtering. The best metric values are highlighted with a star $\star$.}
    \label{fig:tau}
\end{figure*}

\subsubsection{Performance of Aspect Category Sentiment Analysis}

On the Rest15 dataset, our method without data augmentation, Ours (w/o DA) achieves the best F1 score of 77.44\%, which is +9.86\% F1 higher than PBJM (67.58\%). Although the performance of our DA-enhanced model is slightly lower with a F1 score of 74.48\%, it still outperforms PBJM by +6.90\% F1. Although data augmentation does not further improve the performance in this dataset, it maintains competitive results. Similar to Rest15, on the Lap16 dataset, our model without data augmentation (61.56\% F1) performs better than the DA-enhanced version (60.45\% F1). However, these two variants were +5.94\% and +4.83\% F1 higher than PBJM (55.62\% F1), respectively.

Compared with Rest15 and Lap16, our DA-enhanced model achieves the highest F1 score of 82.65\% on the Rest16 dataset, which is +7.62\% F1 higher than PBJM (75.03\%). The without DA variant also performs well (79.58\% F1), outperforming PBJM by +4.55\% F1. These results indicate that our data augmentation strategy is particularly effective in this domain, continuously improving Precision and Recall, resulting in excellent F1 performance. Our model exhibits similar experimental phenomena on the Lap15 dataset as described above, achieving the best F1 score with data augmentation. Our method achieves an F1 score of 66.08\% without data augmentation, still +3.67\% higher than PBJM.

\subsection{Parameter Analysis}

We analyse the effect of hyperparameters $ \alpha $, $\tau$ and the model size in the ACSA task on four benchmark datasets: Rest15, Rest16, Lap15, and Lap16. The detailed experimental results are illustrated in Figure \ref{fig:alpha} and \ref{fig:tau}. In almost all datasets and metrics, our method outperforms the best baseline model PBJM \cite{PBJM} in terms of precision (P), recall (R), and F1 scores.

\subsubsection{Effect of $ \alpha $}
In the Rest15 dataset, compared to PBJM, which achieves an F1 score of 67.58\%, our method consistently outperforms it across all values of $\alpha $, with the best result being 77.51\% at $\alpha $=0.2. Specifically, Precision is highest at $\alpha $=0.2 (79.31\%), and Recall also reaches its peak at $\alpha $=0.2 (75.79\%). Performance steadily declines beyond $\alpha $=0.4, suggesting that increasing the influence of confidence (high $\alpha $) may overemphasise correct predictions at the expense of generalisation. This implies that a lower $\alpha $ value (e.g., 0.2–0.4) is optimal for Rest15 dataset.

In the Rest16 dataset, PBJM achieves an F1 score of 75.03\%, while our method reaches 82.45\% at $\alpha $=1.0, surpassing all other methods. Recall is maximised at $\alpha $=1.0 (83.22\%). This dataset benefits from a larger confidence weight.

In the Lap15 dataset, PBJM performed poorly with an F1 of 62.41\%, while our method improves it to 70.44\% at $\alpha $=0.8. The model's performance steadily improves as $\alpha$ reaches 0.8, followed by a slight decrease at $\alpha$=1.0, with the optimal $\alpha$ range between 0.6 and 0.8.

The most challenging dataset is the Lap16 dataset. PBJM only achieves an F1 score of 55.62\%. Our model improves performance to 60.45\% when $\alpha $=0.4, but the Recall and Precision are relatively low. Our model achieves optimal performance when $\alpha $=0.4, and both accuracy and recall decrease beyond this value.

\input{table/size}

\subsubsection{Effect of $ \tau $}
\label{tau}
To investigate the effect of semantic consistency filtering in data augmentation, we conduct experiments using different thresholds $\tau \in \{0, 0.6, 0.7, 0.8\}$. The parameter $\tau $ controls the post-processing based on the semantic similarity between the generated sentence and the original sentence. Specifically, we retain sentences with cosine similarity above the threshold $\tau $. $\tau=0 $ means to use all generated sentences without any filtering.

It can be observed in Figure \ref{fig:tau} that our method consistently outperforms the PBJM baseline on all metrics and datasets, demonstrating the effectiveness of our augmentation framework. Notably, on Rest16,Lap15, and Lap16, our models trained with filtered augmented data ($\tau>0 $) outperform the unfiltered variant ($\tau=0 $), indicating that removing semantically inconsistent samples can lead to higher quality training data.

On the Rest15 dataset, our model with $\tau=0 $ performs the best, while further filtering slightly reduces performance. This suggests that in some domains, aggressive filtering may discard useful diversity in augmented samples. Nevertheless, even in this case, all of our models with $\tau>0 $ still outperform the PBJM baseline.

Our analysis indicates that a moderate threshold of $\tau=0.7 $ provides the best trade-off and demonstrates superior performance across all datasets.

\begin{figure}[htbp]
    \centering
    \includegraphics[width=0.50\textwidth]{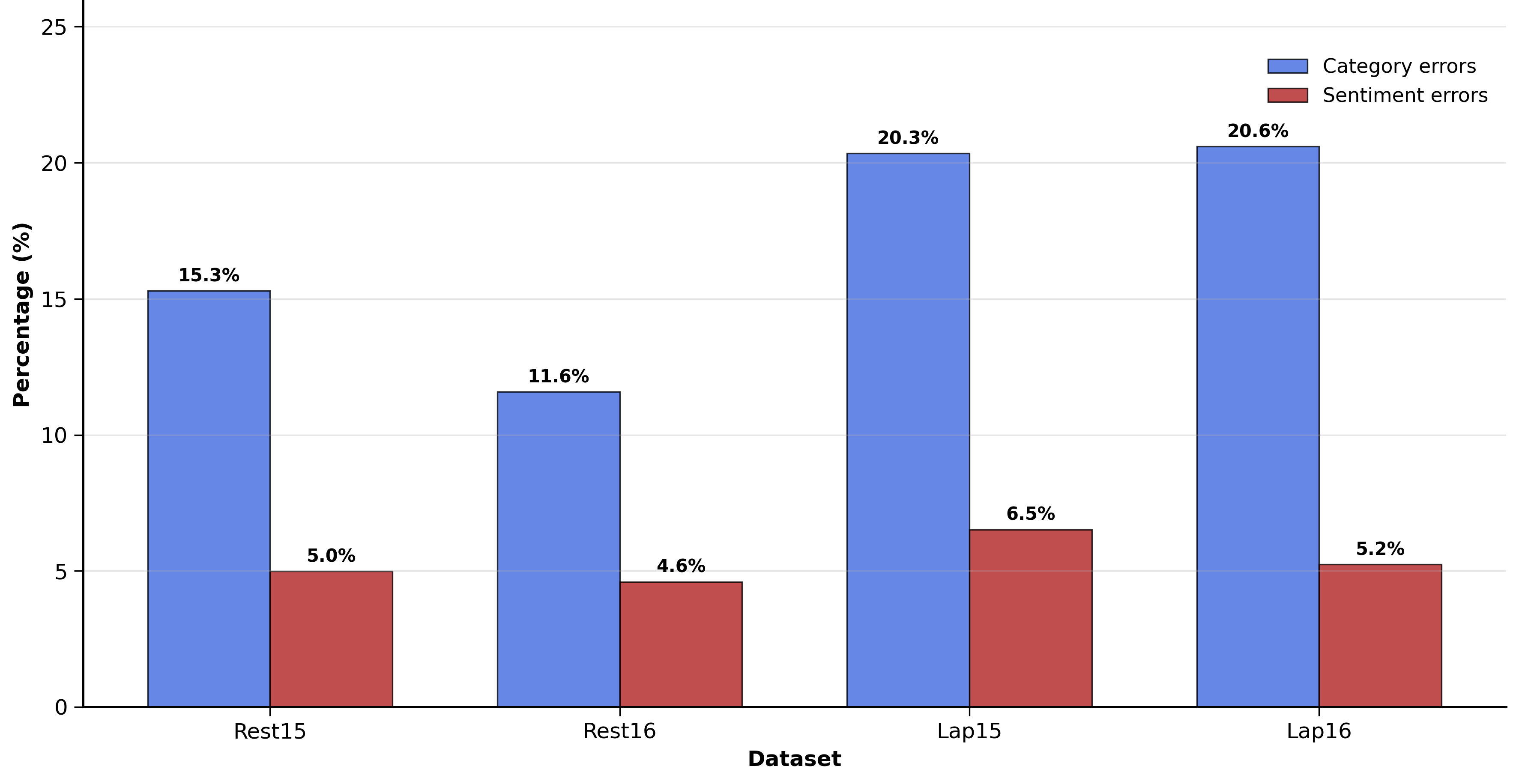}
    \caption{Category error distribution and Sentiment error distribution, which represent predicting sentiment correctly but category wrongly, and predicting category correctly but sentiment wrongly, respectively.}
    \label{fig:C_S}
\end{figure}

\begin{figure*}[htbp]
    \centering
    \includegraphics[width=0.80\textwidth]{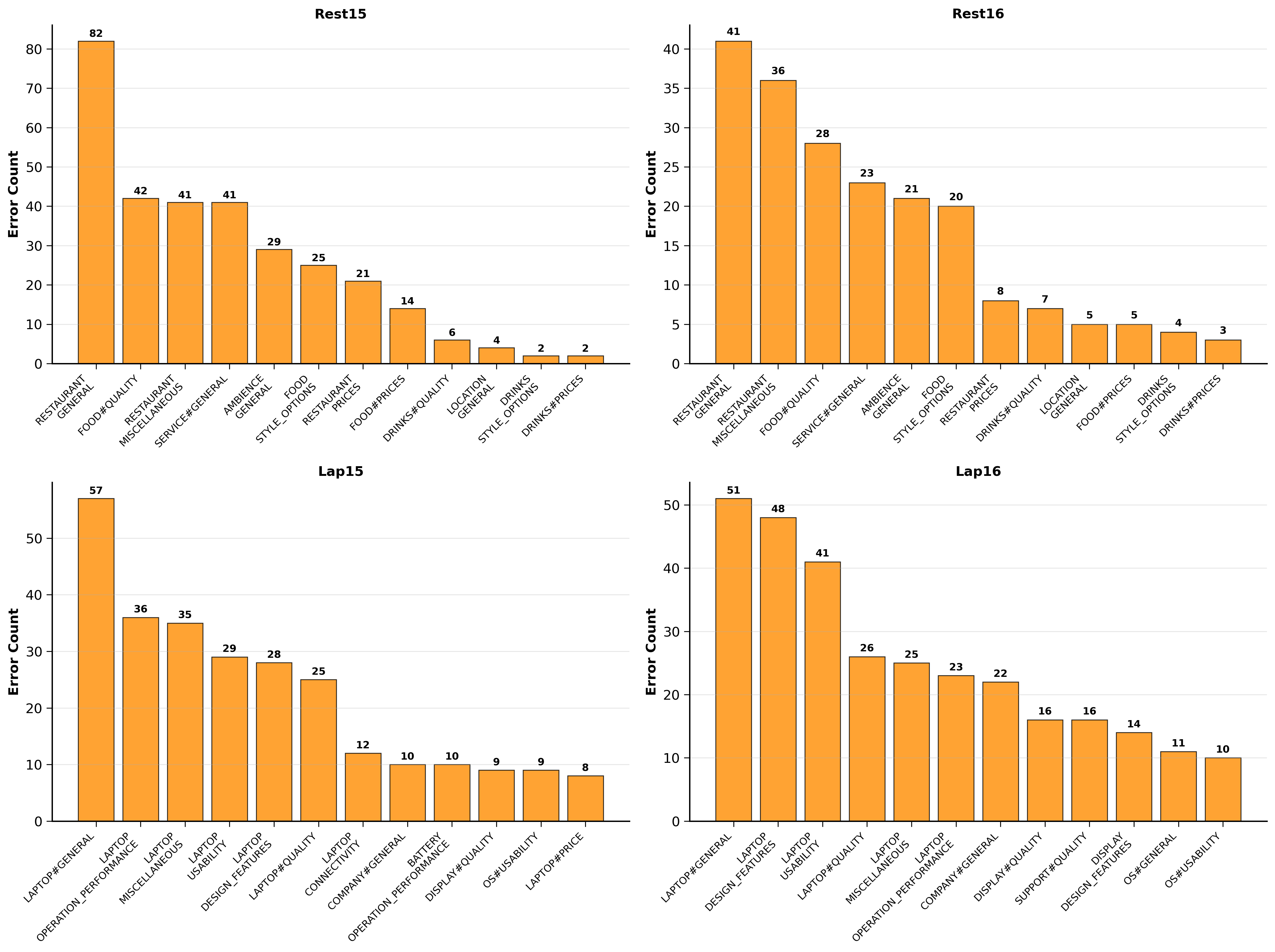}
    \caption{Error counts for mispredicted aspect categories across the four benchmark datasets.}
    \label{fig:top}
\end{figure*}

\subsubsection{Effect of model size}

To investigate the impact of model size on performance, we conduct experiments using our approach based on Flan-T5 with three different model sizes: 220M, 770M, and 3B models. We take Rest16 and Lap16 as examples, and the results for these two datasets are shown in Table \ref{tab:size}.

As the model size increases, we observe consistent improvements in all evaluation metrics. On the Rest16 dataset, the F1 score increases from 56.12\% for the 220M model to 64.12\% for the 770M model, and then to 82.65\% for the 3B model. Lap16 dataset has a similar trend, where the F1 score improves with increasing model size.

The results clearly indicate that expanding the model size can significantly improve performance. Larger models are more effective in capturing complex semantics in aspect category sentiment analysis.

\subsection{Error Analysis}

ACSA is an end-to-end task that requires predicting both category and sentiment correctly to be considered a truly accurate prediction. We analyse the distribution of error samples on four SemEval benchmarks for correctly predicting sentiment but incorrectly predicting category, and correctly predicting category but incorrectly predicting sentiment, as shown in Figure \ref{fig:C_S}. 
\subsubsection{Identifying the correct category is more difficult than identifying its sentiment.}
 Category errors exceed Sentiment errors in all benchmark datasets, and for the Restaurant dataset, Category errors are almost three times that of Sentiment errors. For the Laptop dataset, Category errors rise to four times that of Sentiment errors. This indicates that even in a domain with limited and clear categories, such as FOOD\#QUALITY, LOCATION\#GENERAL and RESTAURANT\#PRICES, when multiple aspect categories appear simultaneously in a sentence, it is still challenging to identify correct categories and predict their sentiments.

\subsubsection{More difficult domains have higher Category error rate.} From the Restaurant to Laptop domain increases the Category error rate by 5\%-9\%, while the Sentiment error rate only slightly increases. As shown in Table \ref{tab:dataset}, the Laptop dataset has increased the number of fine-grained categories and has higher semantic overlap between categories, such a BATTERY\#GENERAL, BATTERY\#MISCELLANEOUS and BATTERY\#OPERATION\_PERFORMANCE. In contrast, there is less ambiguity in the categories of Restaurant dataset, such as FOOD\#QUALITY and RESTAURANT\#PRICES. The richer and more fine-grained categories increase the challenge for the model in predicting aspect categories.

We also conduct a fine-grained analysis of category-level errors on four benchmark datasets. Figure \ref{fig:top} illustrates the distribution of incorrect aspect category predictions for each dataset. From Figure \ref{fig:top}, we can obtain the following observations.
\subsubsection{General categories dominate errors.} The number of RESITAURANT\#GENERAL and LAPTOP\#GENERAL errors always ranks at the top error sources. This underscores that when the model cannot detect fine-grained categories in sentences, these general categories are over-predicted.

\subsubsection{The category-level ambiguity is a major cause of model prediction errors.} Categories with high semantic overlap, such as FOOD\# QUALITY and FOOD\# STYLE\_OPTIONS, or LAPTOP\#DESIGN\_FEATURES and LAPTOP\#USABILITY often appear simultaneously in the error list. This indicates that when the category's semantic expressions are similar, it is challenging for the model to make accurate predictions. Moreover, it is observed that the Laptop dataset exhibits broader semantic overlap between categories. In contrast, in the Restaurant dataset, errors are more concentrated in a few categories.

%% file: table/dataset.tex
\begin{table}[!ht]
\caption{The statistics of each dataset used for the experiment.}
   \label{tab:dataset}
    \centering
\begin{tabular}{llllll}
\toprule
\multicolumn{1}{c}{}                                      & \multicolumn{1}{c}{}      & \multicolumn{1}{c}{Rest15} & \multicolumn{1}{c}{Lap15} & \multicolumn{1}{c}{Rest16} & \multicolumn{1}{c}{Lap16} \\
\midrule
\multicolumn{1}{c}{Training set}                          & \multicolumn{1}{c}{}      & \multicolumn{1}{c}{1120}  & \multicolumn{1}{c}{1399}  & \multicolumn{1}{c}{1708}  & \multicolumn{1}{c}{2039}  \\
\midrule
\multicolumn{1}{c}{Testing set}                           & \multicolumn{1}{c}{}      & \multicolumn{1}{c}{582}   & \multicolumn{1}{c}{644}   & \multicolumn{1}{c}{587}   & \multicolumn{1}{c}{573}   \\
\midrule
\multicolumn{1}{c}{\multirow{2}{*}{Category}} & \multicolumn{1}{c}{Train} & \multicolumn{1}{c}{12} & \multicolumn{1}{c}{12} & \multicolumn{1}{c}{81} & \multicolumn{1}{c}{81} \\
\multicolumn{1}{c}{}                          & \multicolumn{1}{c}{Test}  & \multicolumn{1}{c}{12} & \multicolumn{1}{c}{12} & \multicolumn{1}{c}{58} & \multicolumn{1}{c}{67}\\
\midrule
\multicolumn{1}{c}{\multirow{2}{*}{Positive annotations}} & \multicolumn{1}{c}{Train} & \multicolumn{1}{c}{1198}  & \multicolumn{1}{c}{1103}  & \multicolumn{1}{c}{1657}  & \multicolumn{1}{c}{1637}  \\
\multicolumn{1}{c}{}                                      & \multicolumn{1}{c}{Test}  & \multicolumn{1}{c}{454}   & \multicolumn{1}{c}{541}   & \multicolumn{1}{c}{611}   & \multicolumn{1}{c}{481}   \\
\midrule
\multicolumn{1}{c}{\multirow{2}{*}{Negative annotations}} & \multicolumn{1}{c}{Train} & \multicolumn{1}{c}{403}   & \multicolumn{1}{c}{765}   & \multicolumn{1}{c}{749}   & \multicolumn{1}{c}{1084}  \\
\multicolumn{1}{c}{}                                      & \multicolumn{1}{c}{Test}  & \multicolumn{1}{c}{346}   & \multicolumn{1}{c}{329}   & \multicolumn{1}{c}{204}   & \multicolumn{1}{c}{274}   \\
\midrule
\multicolumn{1}{c}{\multirow{2}{*}{Neutral annotations}}  & \multicolumn{1}{c}{Train} & \multicolumn{1}{c}{53}    & \multicolumn{1}{c}{106}   & \multicolumn{1}{c}{101}   & \multicolumn{1}{c}{188}   \\
\multicolumn{1}{c}{}                                      & \multicolumn{1}{c}{Test}  & \multicolumn{1}{c}{45}    & \multicolumn{1}{c}{79}    & \multicolumn{1}{c}{44}    & \multicolumn{1}{c}{46}     \\
\bottomrule
\end{tabular}
\end{table}

%% file: table/ACSC.tex
\begin{table*}[!ht]
\caption{Experimental results on the ACSC task. $^{\dagger}$ refers that the experimental results are all from ECAN \cite{ecan}. The results marked as \_\_ achieve the best performance among all methods. The scores marked with $^{\ddagger}$ rank second in performance.}
   \label{tab:ACSC_result}
    \centering        

\begin{tabular}{lllllllllllll}
\toprule
\multicolumn{1}{c}{\multirow{2}{*}{Methods}} & \multicolumn{3}{c}{Rest15}                                                                                                                                & \multicolumn{3}{c}{Rest16}                                                                                                                                & \multicolumn{3}{c}{Lap15}                                                                                                                                 & \multicolumn{3}{c}{Lap16}                                                                                                                                 \\
\multicolumn{1}{c}{}                         & \multicolumn{1}{c}{P}                             & \multicolumn{1}{c}{R}                             & \multicolumn{1}{c}{F1}                            & \multicolumn{1}{c}{P}                             & \multicolumn{1}{c}{R}                             & \multicolumn{1}{c}{F1}                            & \multicolumn{1}{c}{P}                             & \multicolumn{1}{c}{R}                             & \multicolumn{1}{c}{F1}                            & \multicolumn{1}{c}{P}                             & \multicolumn{1}{c}{R}                             & \multicolumn{1}{c}{F1}     \\
\midrule
\multicolumn{1}{c}{CAER-BERT $^{\dagger}$}               & \multicolumn{1}{c}{71.27}                         & \multicolumn{1}{c}{63.75}                         & \multicolumn{1}{c}{66.25}                         & \multicolumn{1}{c}{77.69}                         & \multicolumn{1}{c}{68.39}                         & \multicolumn{1}{c}{73.6}                          & \multicolumn{1}{c}{58.34}                         & \multicolumn{1}{c}{51.86}                         & \multicolumn{1}{c}{60.78}                         & \multicolumn{1}{c}{65.36}                         & \multicolumn{1}{c}{54.73}                         & \multicolumn{1}{c}{60.79}                         \\
\multicolumn{1}{c}{SCAN$^{\dagger}$}                    & \multicolumn{1}{c}{71.25}                         & \multicolumn{1}{c}{67.05}                         & \multicolumn{1}{c}{69.09}                         & \multicolumn{1}{c}{73.87}                         & \multicolumn{1}{c}{76.25}                         & \multicolumn{1}{c}{75.05}                         & \multicolumn{1}{c}{71.01}                         & \multicolumn{1}{c}{45.49}                         & \multicolumn{1}{c}{55.45}                         & \multicolumn{1}{c}{61.06}                         & \multicolumn{1}{c}{45.42}                         & \multicolumn{1}{c}{52.08}                         \\
\multicolumn{1}{c}{AC-MIMLLN $^{\dagger}$}               & \multicolumn{1}{c}{71.45}                         & \multicolumn{1}{c}{65.18}                         & \multicolumn{1}{c}{68.17}                         & \multicolumn{1}{c}{75.66}                         & \multicolumn{1}{c}{74.89}                         & \multicolumn{1}{c}{75.27}                         & \multicolumn{1}{c}{59.06}                         & \multicolumn{1}{c}{53.88}                         & \multicolumn{1}{c}{55.29}                         & \multicolumn{1}{c}{68.08}                         & \multicolumn{1}{c}{55.89}                         & \multicolumn{1}{c}{61.39}                         \\
\multicolumn{1}{c}{LC-BERT$^{\dagger}$}                 & \multicolumn{1}{c}{72.08}                         & \multicolumn{1}{c}{65.76}                         & \multicolumn{1}{c}{68.77}                         & \multicolumn{1}{c}{76.34}                         & \multicolumn{1}{c}{74.16}                         & \multicolumn{1}{c}{75.75}                         & \multicolumn{1}{c}{66.84}                         & \multicolumn{1}{c}{50.78}                         & \multicolumn{1}{c}{57.72}                         & \multicolumn{1}{c}{71.05}                         & \multicolumn{1}{c}{53.11}                         & \multicolumn{1}{c}{55.58}                         \\
\multicolumn{1}{c}{EDU-capsule $^{\dagger}$}             & \multicolumn{1}{c}{72.67}                         & \multicolumn{1}{c}{67.34}                         & \multicolumn{1}{c}{69.9}                          & \multicolumn{1}{c}{75.93}                         & \multicolumn{1}{c}{77}                            & \multicolumn{1}{c}{74.81}                         & \multicolumn{1}{c}{73.98}                         & \multicolumn{1}{c}{52.27}                         & \multicolumn{1}{c}{65.43}                         & \multicolumn{1}{c}{67.02}                         & \multicolumn{1}{c}{49.2}                          & \multicolumn{1}{c}{58.01}                         \\
\multicolumn{1}{c}{Hier-BERT $^{\dagger}$}               & \multicolumn{1}{c}{70.42}                         & \multicolumn{1}{c}{65.75}                         & \multicolumn{1}{c}{68.01}                         & \multicolumn{1}{c}{76.99}                         & \multicolumn{1}{c}{73.88}                         & \multicolumn{1}{c}{75.41}                         & \multicolumn{1}{c}{70.19}                         & \multicolumn{1}{c}{48.41}                         & \multicolumn{1}{c}{57.3}                          & \multicolumn{1}{c}{66.01}                         & \multicolumn{1}{c}{53.71}                         & \multicolumn{1}{c}{59.23}                         \\
\multicolumn{1}{c}{Hier-GCN-BERT $^{\dagger}$}           & \multicolumn{1}{c}{75.12}                         & \multicolumn{1}{c}{66.47}                         & \multicolumn{1}{c}{71.94}                         & \multicolumn{1}{c}{77.62}                         & \multicolumn{1}{c}{75.61}                         & \multicolumn{1}{c}{77.68}                         & \multicolumn{1}{c}{75}                            & \multicolumn{1}{c}{66.84}                         & \multicolumn{1}{c}{70.69}                         & \multicolumn{1}{c}{68.32}                         & \multicolumn{1}{c}{62.54}                         & \multicolumn{1}{c}{65.09}                         \\
\multicolumn{1}{c}{MvP $^{\dagger}$}                     & \multicolumn{1}{c}{67.8}                          & \multicolumn{1}{c}{68.63}                         & \multicolumn{1}{c}{68.21}                         & \multicolumn{1}{c}{73.76}                         & \multicolumn{1}{c}{75.49}                         & \multicolumn{1}{c}{74.62}                         & \multicolumn{1}{c}{-}                             & \multicolumn{1}{c}{-}                             & \multicolumn{1}{c}{-}                             & \multicolumn{1}{c}{-}                             & \multicolumn{1}{c}{-}                             & \multicolumn{1}{c}{-}                             \\
\multicolumn{1}{c}{\textbf{ECAN $^{\dagger}$}}           & \multicolumn{1}{c}{84.38}                         & \multicolumn{1}{c}{74.43}                         & \multicolumn{1}{c}{79.09}                         & \multicolumn{1}{c}{84.92}                         & \multicolumn{1}{c}{79.65}                         & \multicolumn{1}{c}{82.2}                          & \multicolumn{1}{c}{83.56}                         & \multicolumn{1}{c}{75.34}                         & \multicolumn{1}{c}{79.24}                         & \multicolumn{1}{c}{76.11}                         & \multicolumn{1}{c}{65.45}                         & \multicolumn{1}{c}{70.45}                         \\
\midrule
\multicolumn{1}{c}{\textbf{Ours (w/o DA)}} & \multicolumn{1}{c}{{\underline{ {\textbf{89.75}}}}} & \multicolumn{1}{c}{{\underline{ {\textbf{90.53}}}}} & \multicolumn{1}{c}{{\underline{ {\textbf{89.46}}}} }& \multicolumn{1}{c}{92.41 $^{\ddagger}$} & \multicolumn{1}{c}{92.67 $^{\ddagger}$} & \multicolumn{1}{c}{91.88 $^{\ddagger}$} & \multicolumn{1}{c}{90.81 $^{\ddagger}$} & \multicolumn{1}{c}{91.46 $^{\ddagger}$} & \multicolumn{1}{c}{90.74 $^{\ddagger}$} & \multicolumn{1}{c}{88.97 $^{\ddagger}$} & \multicolumn{1}{c}{{\underline{ {\textbf{89.39}}}} }& \multicolumn{1}{c}{89.17 $^{\ddagger}$} \\

\multicolumn{1}{c}{\textbf{Ours (DA)}}            & \multicolumn{1}{c}{86.79 $^{\ddagger}$}& \multicolumn{1}{c}{88.28 $^{\ddagger}$ }& \multicolumn{1}{c}{86.72 $^{\ddagger}$}& \multicolumn{1}{c}{{ \underline{ {\textbf{93.84}}}}} & \multicolumn{1}{c}{{\underline{ {\textbf{93.83}}}}} & \multicolumn{1}{c}{{ \underline{ {\textbf{93.44}}}} }& \multicolumn{1}{c}{{\underline{ {\textbf{92.92}}}}} & \multicolumn{1}{c}{{ \underline{ {\textbf{93.36}}}}} & \multicolumn{1}{c}{{ \underline{ {\textbf{92.99}}}}} & \multicolumn{1}{c}{{ \underline{ {\textbf{90.27}}}} }& \multicolumn{1}{c}{89.14 $^{\ddagger}$}& \multicolumn{1}{c}{{ \underline{ {\textbf{89.64}}}}}\\

\bottomrule
\end{tabular}
\end{table*}

%% file: table/ACSA.tex
\begin{table*}[!ht]
\caption{Experimental results on the ACSA task. $^{\ast}$ refers that the experimental results are all from PBJM \cite{PBJM}.  The results marked as \_\_ achieve the best performance among all methods. The scores marked with $^{\ddagger}$ rank second in performance.}
   \label{tab:ACSA_result}
    \centering        
\begin{tabular}{lllllllllllll}
\toprule
\multicolumn{1}{c}{\multirow{2}{*}{Model}} & \multicolumn{3}{c}{Rest15}                                                                                                                                & \multicolumn{3}{c}{Rest16}                                                                                                                                & \multicolumn{3}{c}{Lap15}                                                                                                                                & \multicolumn{3}{c}{Lap16}                                                                                                                                 \\
\multicolumn{1}{c}{}                       & \multicolumn{1}{c}{P}                             & \multicolumn{1}{c}{R}                             & \multicolumn{1}{c}{F1}                            & \multicolumn{1}{c}{P}                             & \multicolumn{1}{c}{R}                             & \multicolumn{1}{c}{F1}                            & \multicolumn{1}{c}{P}                             & \multicolumn{1}{c}{R}                            & \multicolumn{1}{c}{F1}                            & \multicolumn{1}{c}{P}                             & \multicolumn{1}{c}{R}                             & \multicolumn{1}{c}{F1} \\
\midrule
\multicolumn{1}{c}{Pipeline-BERT$^{\ast}$}        & \multicolumn{1}{c}{38.12}                         & \multicolumn{1}{c}{70}                            & \multicolumn{1}{c}{49.35}                         & \multicolumn{1}{c}{43.62}                         & \multicolumn{1}{c}{79.06}                         & \multicolumn{1}{c}{56.21}                         & \multicolumn{1}{c}{36.91}                         & \multicolumn{1}{c}{51.62}                        & \multicolumn{1}{c}{43.02}                         & \multicolumn{1}{c}{31.92}                         & \multicolumn{1}{c}{51.56}                         & \multicolumn{1}{c}{39.42}                         \\
\multicolumn{1}{c}{Cartesian-BERT$^{\ast}$}       & \multicolumn{1}{c}{72.02}                         & \multicolumn{1}{c}{49.15}                         & \multicolumn{1}{c}{58.42}                         & \multicolumn{1}{c}{74.96}                         & \multicolumn{1}{c}{63.84}                         & \multicolumn{1}{c}{68.94}                         & \multicolumn{1}{c}{73.06}                         & \multicolumn{1}{c}{21.18}                        & \multicolumn{1}{c}{32.83}                         & \multicolumn{1}{c}{64.99}                         & \multicolumn{1}{c}{27.4}                          & \multicolumn{1}{c}{39.54}                         \\
\multicolumn{1}{c}{Addonedim-BERT$^{\ast}$}       & \multicolumn{1}{c}{68.84}                         & \multicolumn{1}{c}{55.86}                         & \multicolumn{1}{c}{61.67}                         & \multicolumn{1}{c}{71.75}                         & \multicolumn{1}{c}{67.95}                         & \multicolumn{1}{c}{69.79}                         & \multicolumn{1}{c}{64.13}                         & \multicolumn{1}{c}{39.57}                        & \multicolumn{1}{c}{48.94}                         & \multicolumn{1}{c}{58.83}                         & \multicolumn{1}{c}{39.49}                         & \multicolumn{1}{c}{47.23}                         \\
\multicolumn{1}{c}{AS-DATJM$^{\ast}$}             & \multicolumn{1}{c}{66.35}                         & \multicolumn{1}{c}{50.52}                         & \multicolumn{1}{c}{57.21}                         & \multicolumn{1}{c}{70.88}                         & \multicolumn{1}{c}{60.35}                         & \multicolumn{1}{c}{65.19}                         & \multicolumn{1}{c}{58.91}                         & \multicolumn{1}{c}{40.28}                        & \multicolumn{1}{c}{47.76}                         & \multicolumn{1}{c}{57.29}                         & \multicolumn{1}{c}{36.7}                          & \multicolumn{1}{c}{44.71}                         \\
\multicolumn{1}{c}{Hier-BERT$^{\ast}$}            & \multicolumn{1}{c}{67.46}                         & \multicolumn{1}{c}{57.98}                         & \multicolumn{1}{c}{62.36}                         & \multicolumn{1}{c}{70.97}                         & \multicolumn{1}{c}{69.65}                         & \multicolumn{1}{c}{70.3}                          & \multicolumn{1}{c}{65.47}                         & \multicolumn{1}{c}{41.26}                        & \multicolumn{1}{c}{50.61}                         & \multicolumn{1}{c}{59.51}                         & \multicolumn{1}{c}{41.93}                         & \multicolumn{1}{c}{49.19}                         \\
\multicolumn{1}{c}{Hier-Trans-BERT$^{\ast}$}      & \multicolumn{1}{c}{70.22}                         & \multicolumn{1}{c}{59.96}                         & \multicolumn{1}{c}{64.67}                         & \multicolumn{1}{c}{73.25}                         & \multicolumn{1}{c}{73.21}                         & \multicolumn{1}{c}{73.45}                         & \multicolumn{1}{c}{65.63}                         & \multicolumn{1}{c}{51.95}                        & \multicolumn{1}{c}{57.79}                         & \multicolumn{1}{c}{58.06}                         & \multicolumn{1}{c}{48.29}                         & \multicolumn{1}{c}{52.52}                         \\
\multicolumn{1}{c}{Hier-GCN-BERT$^{\ast}$}        & \multicolumn{1}{c}{71.93}                         & \multicolumn{1}{c}{58.03}                         & \multicolumn{1}{c}{64.23}                         & \multicolumn{1}{c}{76.37}                         & \multicolumn{1}{c}{72.83}                         & \multicolumn{1}{c}{74.55}                         & \multicolumn{1}{c}{71.9}                          & \multicolumn{1}{c}{54.73}                        & \multicolumn{1}{c}{62.13}                         & \multicolumn{1}{c}{61.43}                         & \multicolumn{1}{c}{48.42}                         & \multicolumn{1}{c}{54.15}                         \\
\multicolumn{1}{c}{PBJM$^{\ast}$}                 & \multicolumn{1}{c}{75.07}                         & \multicolumn{1}{c}{61.48}                         & \multicolumn{1}{c}{67.58}                         & \multicolumn{1}{c}{76.53}                         & \multicolumn{1}{c}{73.6}                          & \multicolumn{1}{c}{75.03}                         & \multicolumn{1}{c}{72.23 $^{\ddagger}$}                         & \multicolumn{1}{c}{54.96}                        & \multicolumn{1}{c}{62.41}                         & \multicolumn{1}{c}{62.63}                         & \multicolumn{1}{c}{50.08}                         & \multicolumn{1}{c}{55.62}                         \\
\midrule
\multicolumn{1}{c}{\textbf{Ours (w/o DA)}} & \multicolumn{1}{c}{{\underline{ {\textbf{79.73}}}} }& \multicolumn{1}{c}{{\underline{ {\textbf{75.29}}}} }& \multicolumn{1}{c}{{\underline{ {\textbf{77.44}}}}} & \multicolumn{1}{c}{79.26 $^{\ddagger}$}                         & \multicolumn{1}{c}{79.89 $^{\ddagger}$}                         & \multicolumn{1}{c}{79.58 $^{\ddagger}$}                         & \multicolumn{1}{c}{68.96 }                         & \multicolumn{1}{c}{63.44 $^{\ddagger}$}                        & \multicolumn{1}{c}{66.08 $^{\ddagger}$}                         & \multicolumn{1}{c}{{\underline{ {\textbf{64.14}}}}} & \multicolumn{1}{c}{{\underline{ {\textbf{59.18}}}}}& \multicolumn{1}{c}{{\underline{ {\textbf{61.56}}}}} \\
\multicolumn{1}{c}{\textbf{Ours (DA)}}          & \multicolumn{1}{c}{76.16 $^{\ddagger}$}                         & \multicolumn{1}{c}{72.88 $^{\ddagger}$}                         & \multicolumn{1}{c}{74.48 $^{\ddagger}$}                         & \multicolumn{1}{c}{{\underline{ {\textbf{81.58}}}} }& \multicolumn{1}{c}{{\underline{ {\textbf{83.75}}}}} & \multicolumn{1}{c}{{\underline{ {\textbf{82.65}}}} }& \multicolumn{1}{c}{{\underline{ {\textbf{72.74}}}} }& \multicolumn{1}{c}{{\underline{ {\textbf{68.6}}}}} & \multicolumn{1}{c}{{\underline{ {\textbf{70.61}}}}}& \multicolumn{1}{c}{62.77 $^{\ddagger}$}                         & \multicolumn{1}{c}{58.30 $^{\ddagger}$}                         & \multicolumn{1}{c}{60.45 $^{\ddagger}$}                         \\
\bottomrule
\end{tabular}
\end{table*}

%% file: table/size.tex
\begin{table}[]
\caption{Evaluation results of our methods under different model sizes for the ACSA task.}
   \label{tab:size}
\begin{tabular}{lllllll}
\toprule
\multicolumn{1}{c}{\multirow{2}{*}{Model}} & \multicolumn{3}{c}{Rest16}                                                        & \multicolumn{3}{c}{Lap16}                                                         \\
\multicolumn{1}{c}{}                       & \multicolumn{1}{c}{P}     & \multicolumn{1}{c}{R}     & \multicolumn{1}{c}{F1}    & \multicolumn{1}{c}{P}     & \multicolumn{1}{c}{R}     & \multicolumn{1}{c}{F1}    \\
\midrule
\multicolumn{1}{c}{Ours (220M)}            & \multicolumn{1}{c}{63.42} & \multicolumn{1}{c}{50.33} & \multicolumn{1}{c}{56.12} & \multicolumn{1}{c}{36.38} & \multicolumn{1}{c}{23.85} & \multicolumn{1}{c}{28.81} \\
\multicolumn{1}{c}{Ours (770M)}            & \multicolumn{1}{c}{68.43} & \multicolumn{1}{c}{60.32} & \multicolumn{1}{c}{64.12} & \multicolumn{1}{c}{60.18} & \multicolumn{1}{c}{50.56} & \multicolumn{1}{c}{54.95} \\
\multicolumn{1}{c}{Ours (3B)}              & \multicolumn{1}{c}{81.58} & \multicolumn{1}{c}{83.75} & \multicolumn{1}{c}{82.65} & \multicolumn{1}{c}{62.77} & \multicolumn{1}{c}{58.3}  & \multicolumn{1}{c}{60.45} \\
\bottomrule
\end{tabular}
\end{table}

%% file: 5_conclusion.tex
\section{Conclusion}

We introduce a data augmentation method that utilises the large language model to generate a dataset that is semantically consistent with the original sentence and demonstrates linguistic diversity. To further ensure data quality, we introduce a post-processing technique to filter generated sentences that deviate too much from the original sentence semantics, aiming to solve the data scarcity challenge in the ACSA task. Fine-tuning enables pre-trained language models to adapt to specific domain sentiment patterns. It enables the model to capture the correlation between aspect categories and sentiment polarity in the target dataset. To enable the model to better adapt to the ACSA task, we propose a confidence-weighted fine-tuning strategy to train the model, encouraging it to learn high confidence and correct predictions. Our experiments on four publicly available datasets are all superior to the baseline models, effectively improving the existing performance of the ACSA task.

%% file: main.bbl
\begin{thebibliography}{10}
\providecommand{\url}[1]{#1}
\csname url@samestyle\endcsname
\providecommand{\newblock}{\relax}
\providecommand{\bibinfo}[2]{#2}
\providecommand{\BIBentrySTDinterwordspacing}{\spaceskip=0pt\relax}
\providecommand{\BIBentryALTinterwordstretchfactor}{4}
\providecommand{\BIBentryALTinterwordspacing}{\spaceskip=\fontdimen2\font plus
\BIBentryALTinterwordstretchfactor\fontdimen3\font minus \fontdimen4\font\relax}
\providecommand{\BIBforeignlanguage}[2]{{%
\expandafter\ifx\csname l@#1\endcsname\relax
\typeout{** WARNING: IEEEtran.bst: No hyphenation pattern has been}%
\typeout{** loaded for the language `#1'. Using the pattern for}%
\typeout{** the default language instead.}%
\else
\language=\csname l@#1\endcsname
\fi
#2}}
\providecommand{\BIBdecl}{\relax}
\BIBdecl

\bibitem{DL-absa}
X.~Chen, H.~Xie, S.~J. Qin, Y.~Chai, X.~Tao, and F.~L. Wang, ``Cognitive-inspired deep learning models for aspect-based sentiment analysis: A retrospective overview and bibliometric analysis,'' \emph{Cogn. Comput.}, vol.~16, no.~6, pp. 3518--3556, 2024.

\bibitem{add1}
L.~Xiao, R.~Mao, S.~Zhao, Q.~Lin, Y.~Jia, L.~He, and E.~Cambria, ``Exploring cognitive and aesthetic causality for multimodal aspect-based sentiment analysis,'' \emph{IEEE Trans. Affective Comput.}, pp. 1--18, 2025.

\bibitem{add2}
Z.~Lian, R.~Liu, K.~Xu, B.~Liu, X.~Liu, Y.~Zhang, X.~Liu, Y.~Li, Z.~Cheng, H.~Zuo, Z.~Ma, X.~Peng, X.~Chen, Y.~Li, E.~Cambria, G.~Zhao, B.~W. Schuller, and J.~Tao, ``Mer 2025: When affective computing meets large language models,'' \emph{arXiv preprint arXiv:2504.19423}, 2025.

\bibitem{cmlm}
Q.~Cheng, J.~Huang, and Y.~Duan, ``Semantically consistent data augmentation for neural machine translation via conditional masked language model,'' in \emph{Proc. Int. Conf. Comput. Linguist. (COLING)}.\hskip 1em plus 0.5em minus 0.4em\relax Int. Comm. Comput. Linguistics, 2022, pp. 5148--5157.

\bibitem{SPDAug-ABSA}
T.~Hsu, C.~Chen, H.~Huang, and H.~Chen, ``Semantics-preserved data augmentation for aspect-based sentiment analysis,'' in \emph{Proc. Conf. Empirical Methods Nat. Lang. Process. (EMNLP)}.\hskip 1em plus 0.5em minus 0.4em\relax Assoc. Comput. Linguistics, 2021, pp. 4417--4422.

\bibitem{rsda}
H.~Wang, K.~He, B.~Li, L.~Chen, F.~Li, X.~Han, C.~Teng, and D.~Ji, ``Refining and synthesis: A simple yet effective data augmentation framework for cross-domain aspect-based sentiment analysis,'' in \emph{Findings of the Association for Computational Linguistics (ACL)}.\hskip 1em plus 0.5em minus 0.4em\relax Assoc. Comput. Linguistics, 2024, pp. 10\,318--10\,329.

\bibitem{DS2-ABSA}
H.~Xu, Y.~Zhang, Q.~Wang, and R.~Xu, ``Ds$^2$-absa: Dual-stream data synthesis with label refinement for few-shot aspect-based sentiment analysis,'' \emph{arXiv preprint arXiv:2412.14849}, 2024.

\bibitem{AAGCN}
B.~Liang, H.~Su, R.~Yin, L.~Gui, M.~Yang, Q.~Zhao, X.~Yu, and R.~Xu, ``Beta distribution guided aspect-aware graph for aspect category sentiment analysis with affective knowledge,'' in \emph{Proc. Conf. Empirical Methods Nat. Lang. Process. (EMNLP)}, 2021, pp. 208--218.

\bibitem{AC-MIMLLN}
Y.~Li, C.~Yin, S.~Zhong, and X.~Pan, ``Multi-instance multi-label learning networks for aspect-category sentiment analysis,'' in \emph{Proc. Conf. Empirical Methods Nat. Lang. Process. (EMNLP)}, 2020, pp. 3550--3560.

\bibitem{PBJM}
Z.~Ping, G.~Sang, Z.~Liu, and Y.~Zhang, ``Aspect category sentiment analysis based on prompt-based learning with attention mechanism,'' \emph{Neurocomputing}, vol. 565, p. 126994, 2024.

\bibitem{ACSA-gen}
J.~Liu, Z.~Teng, L.~Cui, H.~Liu, and Y.~Zhang, ``Solving aspect category sentiment analysis as a text generation task,'' in \emph{Proc. Conf. Empirical Methods Nat. Lang. Process. (EMNLP)}, 2021, pp. 4406--4416.

\bibitem{bart}
M.~Lewis, Y.~Liu, N.~Goyal, M.~Ghazvininejad, A.~Mohamed, O.~Levy, V.~Stoyanov, and L.~Zettlemoyer, ``Bart: Denoising sequence-to-sequence pre-training for natural language generation, translation, and comprehension,'' in \emph{Proc. Annu. Meeting Assoc. Comput. Linguistics (ACL)}.\hskip 1em plus 0.5em minus 0.4em\relax Assoc. Comput. Linguistics, 2020, pp. 7871--7880.

\bibitem{pyacsa}
L.~D. Quilio and F.~Fioravanti, ``A comprehensive framework for aspect-category sentiment analysis,'' in \emph{Proc. 8th Workshop Natural Lang. Artif. Intell. (NL4AI) at Int. Conf. Italian Assoc. Artif. Intell. (AI*IA)}, 2024.

\bibitem{flan-t5}
H.~W. Chung, L.~Hou, S.~Longpre, B.~Zoph, Y.~Tay, W.~Fedus, Y.~Li, X.~Wang, M.~Dehghani, S.~Brahma \emph{et~al.}, ``Scaling instruction-finetuned language models,'' \emph{J. Mach. Learn. Res.}, vol.~25, no.~70, pp. 1--53, 2024.

\bibitem{LEGO-ABSA}
T.~Gao, J.~Fang, H.~Liu, Z.~Liu, C.~Liu, P.~Liu, Y.~Bao, and W.~Yan, ``Lego-absa: A prompt-based task assemblable unified generative framework for multi-task aspect-based sentiment analysis,'' in \emph{Proc. Int. Conf. Comput. Linguistics (COLING)}, 2022, pp. 7002--7012.

\bibitem{chatabsa}
Y.~Bai, Z.~Han, Y.~Zhao, H.~Gao, Z.~Zhang, X.~Wang, and M.~Hu, ``Is compound aspect-based sentiment analysis addressed by llms?'' in \emph{Findings Assoc. Comput. Linguistics: EMNLP}, 2024, pp. 7836--7861.

\bibitem{BERT}
J.~Devlin, M.~Chang, K.~Lee, and K.~Toutanova, ``Bert: Pre-training of deep bidirectional transformers for language understanding,'' in \emph{Proc. Conf. North Amer. Chapter Assoc. Comput. Linguistics: Human Lang. Technol. (NAACL-HLT)}.\hskip 1em plus 0.5em minus 0.4em\relax Assoc. Comput. Linguistics, 2019, pp. 4171--4186.

\bibitem{gpt-3}
T.~B. Brown, B.~Mann, N.~Ryder, M.~Subbiah, J.~Kaplan, P.~Dhariwal, A.~Neelakantan, P.~Shyam, G.~Sastry, A.~Askell, S.~Agarwal, A.~Herbert-Voss, G.~Krueger, T.~Henighan, R.~Child, A.~Ramesh, D.~M. Ziegler, J.~Wu, C.~Winter, C.~Hesse, M.~Chen, E.~Sigler, M.~Litwin, S.~Gray, B.~Chess, J.~Clark, C.~Berner, S.~McCandlish, A.~Radford, I.~Sutskever, and D.~Amodei, ``Language models are few-shot learners,'' in \emph{Adv. Neural Inf. Process. Syst.}, vol.~33, 2020.

\bibitem{cot}
J.~Wei, X.~Wang, D.~Schuurmans, M.~Bosma, B.~Ichter, F.~Xia, E.~H. Chi, Q.~V. Le, and D.~Zhou, ``Chain-of-thought prompting elicits reasoning in large language models,'' in \emph{Adv. Neural Inf. Process. Syst.}, vol.~35, 2022.

\bibitem{self-consistency}
X.~Wang, J.~Wei, D.~Schuurmans, Q.~V. Le, E.~H. Chi, S.~Narang, A.~Chowdhery, and D.~Zhou, ``Self-consistency improves chain of thought reasoning in language models,'' in \emph{Proc. Int. Conf. Learn. Represent. (ICLR)}.\hskip 1em plus 0.5em minus 0.4em\relax OpenReview.net, 2023.

\bibitem{rvisa}
W.~Lai, H.~Xie, G.~Xu, and Q.~Li, ``Rvisa: Reasoning and verification for implicit sentiment analysis,'' \emph{IEEE Trans. Affect. Comput.}, 2025.

\bibitem{da}
Y.~Chai, H.~Xie, and S.~J. Qin, ``Text data augmentation for large language models: A comprehensive survey of methods, challenges, and opportunities,'' \emph{CoRR}, vol. abs/2501.18845, 2025.

\bibitem{sbert}
N.~Reimers and I.~Gurevych, ``Sentence-bert: Sentence embeddings using siamese bert-networks,'' in \emph{Proc. Conf. Empirical Methods Nat. Lang. Process. and Int. Joint Conf. Nat. Lang. Process. (EMNLP-IJCNLP)}.\hskip 1em plus 0.5em minus 0.4em\relax Assoc. Comput. Linguistics, 2019, pp. 3980--3990.

\bibitem{sem15}
M.~Pontiki, D.~Galanis, H.~Papageorgiou, S.~Manandhar, and I.~Androutsopoulos, ``Semeval-2015 task 12: Aspect based sentiment analysis,'' in \emph{Proc. Int. Workshop Semantic Eval. (SemEval@NAACL-HLT)}.\hskip 1em plus 0.5em minus 0.4em\relax Assoc. Comput. Linguistics, 2015, pp. 486--495.

\bibitem{sem16}
M.~Pontiki, D.~Galanis, H.~Papageorgiou, I.~Androutsopoulos, S.~Manandhar, M.~Al{-}Smadi, M.~Al{-}Ayyoub, Y.~Zhao, B.~Qin, O.~D. Clercq, V.~Hoste, M.~Apidianaki, X.~Tannier, N.~V. Loukachevitch, E.~V. Kotelnikov, N.~Bel, S.~M. Jim{\'{e}}nez{-}Zafra, and G.~Eryigit, ``Semeval-2016 task 5: Aspect based sentiment analysis,'' in \emph{Proc. Int. Workshop Semantic Eval. (SemEval@NAACL-HLT)}.\hskip 1em plus 0.5em minus 0.4em\relax Assoc. Comput. Linguistics, 2016, pp. 19--30.

\bibitem{ecan}
J.~Cui, F.~Fukumoto, X.~Wang, Y.~Suzuki, J.~Li, N.~Tomuro, and W.~Kong, ``Enhanced coherence-aware network with hierarchical disentanglement for aspect-category sentiment analysis,'' in \emph{Proc. Joint Int. Conf. Comput. Linguistics, Lang. Resources and Eval. (LREC-COLING)}, 2024, pp. 5843--5855.

\bibitem{CAER-BERT}
B.~Liang, J.~Du, R.~Xu, B.~Li, and H.~Huang, ``Context-aware embedding for targeted aspect-based sentiment analysis,'' in \emph{Proc. Annu. Meeting Assoc. Comput. Linguistics (ACL)}.\hskip 1em plus 0.5em minus 0.4em\relax Assoc. Comput. Linguistics, 2019, pp. 4678--4683.

\bibitem{SCAN}
Y.~Li, C.~Yin, and S.~Zhong, ``Sentence constituent-aware aspect-category sentiment analysis with graph attention networks,'' in \emph{Proc. Int. Conf. Natural Lang. Process. and Chinese Comput. (NLPCC)}, ser. Lecture Notes in Comput. Sci., vol. 12430.\hskip 1em plus 0.5em minus 0.4em\relax Springer, 2020, pp. 815--827.

\bibitem{LC-BERT}
Y.~Wu, Z.~Zhang, Y.~Zhao, and B.~Qin, ``Locate and combine: A two-stage framework for aspect-category sentiment analysis,'' in \emph{Proc. 10th CCF Int. Conf. Natural Lang. Process. Chinese Comput. (NLPCC)}, ser. Lect. Notes Comput. Sci., vol. 13028.\hskip 1em plus 0.5em minus 0.4em\relax Springer, 2021, pp. 595--606.

\bibitem{EDU-Capsule}
T.~Lin, A.~Sun, and Y.~Wang, ``Edu-capsule: Aspect-based sentiment analysis at clause level,'' \emph{Knowl. Inf. Syst.}, vol.~65, no.~2, pp. 517--541, 2023.

\bibitem{Hier-BERT}
H.~Cai, Y.~Tu, X.~Zhou, J.~Yu, and R.~Xia, ``Aspect-category based sentiment analysis with hierarchical graph convolutional network,'' in \emph{Proc. 28th Int. Conf. Comput. Linguistics (COLING)}.\hskip 1em plus 0.5em minus 0.4em\relax Int. Comm. Comput. Linguistics, 2020, pp. 833--843.

\bibitem{mvp}
Z.~Gou, Q.~Guo, and Y.~Yang, ``Mvp: Multi-view prompting improves aspect sentiment tuple prediction,'' in \emph{Proc. 61st Annu. Meeting Assoc. Comput. Linguistics (ACL)}, 2023, pp. 4380--4397.

\bibitem{Addonedim-BERT}
M.~Schmitt, S.~Steinheber, K.~Schreiber, and B.~Roth, ``Joint aspect and polarity classification for aspect-based sentiment analysis with end-to-end neural networks,'' in \emph{Proc. Conf. Empirical Methods Nat. Lang. Process. (EMNLP)}, 2018, pp. 1109--1114.

\bibitem{AS-DATJM}
P.~Gu and Z.~Zhang, ``Dual-attention based joint aspect sentiment classification model,'' in \emph{Int. Conf. Web Eng. (ICWE)}.\hskip 1em plus 0.5em minus 0.4em\relax Springer, 2022, pp. 252--267.

\end{thebibliography}
